%% file: main.tex
\pdfoutput=1 %

\documentclass[11pt]{article}
\usepackage{coling2020}
\usepackage{times}
\usepackage{url}
\usepackage{latexsym}

\usepackage{amsmath,amssymb,mathtools}
\usepackage{enumitem}

\usepackage{pifont}

\usepackage[table,xcdraw]{xcolor}

\setlist{%
itemsep=0pt, parsep=0pt, topsep=0pt, partopsep=0pt, %
leftmargin=*, labelsep=0.5em %
}

\usepackage{microtype}
\usepackage{booktabs}
\usepackage{multirow}

\usepackage{capt-of}
\usepackage{wrapfig}

\newcommand{\sentencedel}{\textrm{SD}}
\newcommand{\verbanonym}{\textrm{VA}}
\newcommand{\pasanonym}{\textrm{PAA}}

\usepackage{color}

\colingfinalcopy %

\newcommand{\propplearner}{\textrm{ProppLearner}}

\title{Modeling Event Salience in Narratives via Barthes' Cardinal Functions}

\author{
    \bf{Takaki Otake}$^{1}$\ \ \ 
    \bf{Sho Yokoi}$^{1,2}$\ \ \ 
    \bf{Naoya Inoue}$^{1,2}$ \thanks{~~~Present affiliation: Stony Brook University.} \\
    \bf{Ryo Takahashi}$^{1,2}$\ \ \ 
    \bf{Tatsuki Kuribayashi}$^{1,3}$\ \ \ 
    \bf{Kentaro Inui}$^{1,2}$ \\
    $^{1}$Tohoku University\ \ \ $^{2}$RIKEN\ \ \ $^{3}$Langsmith Inc. \\
    {\tt \{takaki, yokoi, ryo.t, kuribayashi, inui\}@ecei.tohoku.ac.jp} \\
    {\tt naoya.inoue.lab@gmail.com}
}

\date{}

\begin{document}
\maketitle
\begin{abstract}
Events in a narrative differ in \emph{salience}: some are more important to the story than others.
Estimating event salience is useful for tasks such as story generation, and as a tool for text analysis in narratology and folkloristics.
To compute event salience without any annotations, we adopt Barthes' definition of event salience and propose several unsupervised methods that require only a pre-trained language model.
Evaluating the proposed methods on folktales with event salience annotation, we show that the proposed methods outperform baseline methods and find fine-tuning a language model on narrative texts is a key factor in improving the proposed methods.
\end{abstract}

\section{Introduction}
\label{sec:introduction}
\blfootnote{

    \hspace{-0.65cm}  %
    This work is licensed under a Creative Commons 
    Attribution 4.0 International Licence.
    Licence details:
    \url{http://creativecommons.org/licenses/by/4.0/}.
     
}
Narratives (e.g., folktales, literary short stories) are the representations of a series of events~\cite{abbott2008:introduction}.
Events, the essential components of narratives, differ in \emph{salience}: some are more important to the story than others.
Taking \emph{Cinderella} as an example,  \textit{The prince falls in love with Cinderella} is a salient event; however, \textit{Cinderella draws water from a well} is not.
Estimating event salience is a fundamental task in analyzing and processing narratives, ranging from narrative analysis to automatic story generation~\cite{ouyang2015emnlp:repotable,choubey2018naacl:dominant,papalampidi2019acl:movie}.

This study aims to estimate event salience in an unsupervised manner.
Manually annotating event salience is prohibitively costly because it requires annotators to deeply understand the notion of event salience in narratology~\cite{finlayson2015:ProppLearner}.
In fact, despite a long history of research, very few narrative corpora are annotated with event salience.
Thus, it is crucial to develop a method for estimating event salience that does not rely on event salience-annotated corpora.

In order to estimate event salience without annotated data, we adopt the definition of \emph{cardinal functions} (\textbf{CF}s) introduced by Barthes~\shortcite{barthes1966:introduction_original,barthes1975:introduction}, the successor of \emph{Proppian function}~\footnotemark~\cite{propp1928:morfologiya}, as follows:
\begin{quote}\label{cardinal_function_def}
    \textit{cardinal functions are logically essential to the narrative action and cannot be eliminated without destroying its causal-chronological coherence.}~\cite{prince2003:dictionary}
\end{quote}
\noindent
This definition suggests a simple test for identifying event salience: an event is highly salient if removing it greatly reduces the story's coherence.
We adopt this idea for two reasons. 
First, CFs are commonly used in narrative analysis~\cite{abbott2008:introduction}.
Second, the idea of CFs can be directly operationalized without any annotated data.
Computing event salience based on the idea of CFs requires measuring narrative texts' coherence, but recent advances in discourse coherence models can provide a solution for this difficulty.
To date, a wide variety of discourse coherence models have been proposed~\cite{barzilay2008cl:entitygrid,jurafsky2017emnlp:coherence}.
\newcite{see2019conll:massively} have reported that GPT-2~\cite{radford2019:gpt2}, a powerful left-to-right language model (LM), could accurately estimate narrative texts' coherence, importantly, without any annotated data.
Note that, in folkloristics and narratology, another well-known concept of event salience, \emph{motif} is ``the smallest element in a tale having a power to persist in tradition''~\cite{thompson1946:folktale}, but CFs are more operationalizable given accurate discourse coherence models.

\footnotetext{\emph{Proppian function} is defined as ``an act of character, defined from the point of view of its significance for the course of the action''~\cite{propp1928:morfologiya,propp1968:morphology}}

\section{Related work}
\label{sec:related_work}
Numerous studies on the salience of text units (e.g., word, sentence) can be related to our work.
Here, we review two particularly relevant topics.
First, the deletion test~\cite{carlson2001:discourse} aims to identify salient discourse segments in rhetorical structure theory~\cite{mann1987:rhetorical}.
In the deletion test, annotators check how much discourse coherence is reduced by removing the discourse unit of interest.
Notably,~\newcite{carlson2001:discourse} and ~\newcite{barthes1966:introduction_original} use essentially the same idea, to ``remove the textual unit of interest, and see how the whole structure changes,'' although the task is quite different.
Second, extractive summarization is a task of identifying salient sentences in documents, which is formally very similar to the task of our work.
Despite various existing approaches for extractive summarization~\cite{mani2001:summarization,Gambhir2017:summarization}, it is the open problem whether these methods can be directly applied to narrative texts.
Extractive summarization conventionally focuses on domains with rigid structures, such as news articles or scientific papers, while narrative texts do not have such rigid structures~\cite{kazantseva2010:summarizing}.

In the context of narrative processing in NLP, several methods have been proposed to identify some kinds of salient events: suspenseful events in entertainment stories~\cite{wilmot2020acl:suspense}, turning points in a movie script~\cite{papalampidi2019acl:movie}, and reportable events in personal narratives~\cite{ouyang2015emnlp:repotable}.
In contrast to studies focusing on a specific type of narrative (e.g., movie scripts, personal narratives), our method is potentially applicable to any type of narrative because Barthes' CFs is not a concept specific to those particular kinds of narratives and because our methods require only a pre-trained language model.
\begin{wrapfigure}[10]{r}{0.5\textwidth}
    \centering
        \includegraphics[width=0.48\textwidth]{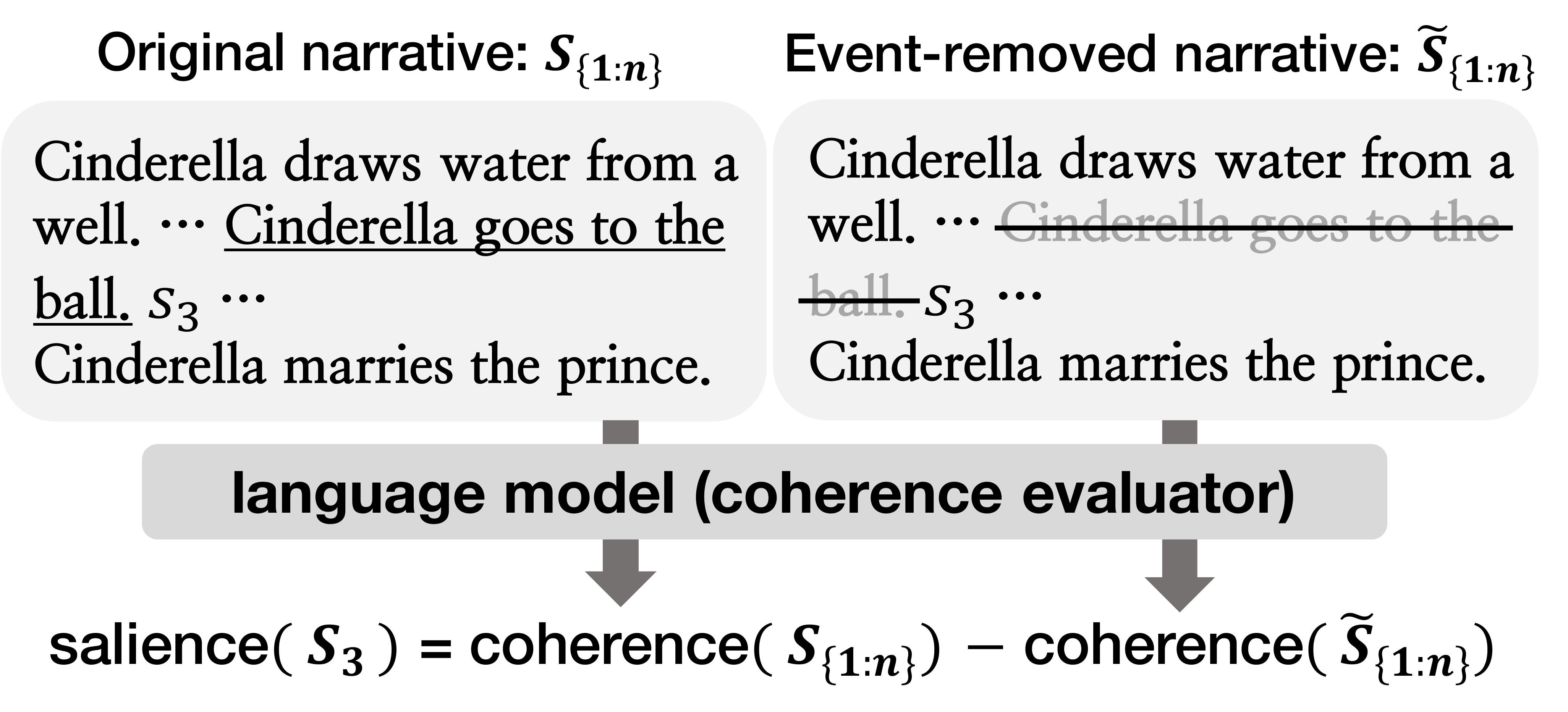} 
        \captionof{figure}{The basic idea of our method based on the definition of Barthes' cardinal function.}
        \label{fig:cardinal-function}
\end{wrapfigure}
\section{Estimating event salience}
\label{sec:estimating_event_salience}
\subsection{Task setup}
\label{task-setup}
We identify event salience in the simplified setting introduced by \newcite{ouyang2015emnlp:repotable}.
That is, we estimate a \textit{sentence's salience} rather than an \textit{event's salience};
we score each sentence in a narrative according to the degree to which it contains a salient event.
This simplification enables us to avoid the difficult subtask of identifying phrases and clauses that express events, while addressing the task of identifying sentences that express salient events. 
Moreover, this sentence level identification can be easily applied to narrative processing and narrative analysis.
Formally, given a narrative comprising $n$ sentences $S_{\{1:n\}} := \{S_1,\dots,S_n\}$ and the target sentence $S_k \in S_{\{1:n\}}$, our goal is to predict the salience score of $S_k$ in $S_{\{1:n\}}$, denoted by $\sigma(S_k, S_{\{1:n\}})\in\mathbb R$.
\subsection{Proposed method}
\label{modeling-event-salience}
\paragraph{Overview}
In light of Barthes' definition (Section~\ref{cardinal_function_def}), we compute the salience score $\sigma(S_k, S_{\{1:n\}})$ as the amount of coherence loss when events in $S_k$ are deleted from the original narrative $S_{\{1:n\}}$ (Figure~\ref{fig:cardinal-function}).
If a narrative's coherence is greatly reduced when events in a sentence are removed, the sentence is considered to contain a highly salient event.

To this end, let $\widetilde{S}_{\{1:n\}} := \{S_{\{1:k-1\}}, r(S_k), S_{\{k+1:n\}}\}$ be the modified narrative with all events in $S_k$ removed from the given narrative $S_{\{1:n\}}$,
where $r$ is an \emph{event removal function}, introduced in the following paragraph.
Let $c(S)$ be the \emph{coherence score} of a given narrative $S$.
Then, the salience score of $S_k$ can be estimated as follows:
\begin{align}
    \sigma(S_k,S_{\{1:n\}}) := c(S_{\{1:n\}}) - c(\widetilde{S}_{\{1:n\}})
\end{align}
In the following, we describe the details of (i) an event removal function $r$ and (ii) coherence evaluator $c$.
\paragraph{Removing events in a sentence: $r$}
\label{subsec:del}
We employ the following three functions $r$. %
\begin{enumerate}
    \item Sentence Deletion (\textbf{\sentencedel{}}): Removing the entire sentence
    \item Verb Anonymization (\textbf{\verbanonym{}}): Replacing all verbs in the sentence with common verbs (e.g., ``do'', ``does'', ``did'') based on the POS tags of each verb~\footnote{We replace verbs with ``do'', ``does'', ``did'', ``done'', or ``doing''. For example, we replace the verb whose POS tag is VBZ (verb, third person singular present) with ``does''.}
    \item Predicate and Arguments Anonymization (\textbf{\pasanonym{}}): Replacing all verbs with common verbs (as in VA) and their main arguments with an indefinite pronoun (e.g., ``someone'', ``something'')~\footnote{We replace ARG0 (i.e., agent) with ``someone'' and ARG1 (i.e., patient) with ``something''.}
\end{enumerate}
We employ \verbanonym{} and \pasanonym{} because predicates and their arguments are main components of commonly used event representations~\cite{chambers-jurafsky2009acl:unsupervised,pichotta2016AAAI:scripts,martin2018AAAI:event,niklaus2018coling:openie}.
\paragraph{Computing narratives' coherence: $c$}
Following~\newcite{see2019conll:massively}, we compute the generation probability of a narrative using a pre-trained language model and regard it as the narrative's coherence score.
Importantly, pre-trained LMs allow us to evaluate narrative's coherence without any annotated data.

Here, the narrative's generation probability is the product of word probabilities, which is influenced by the number of words in the narrative.
Thus, following~\newcite{jurafsky2017emnlp:coherence}, we estimate the coherence score by the \emph{average} log-likelihood of all tokens.
Moreover, we consider only sentences after the target sentence $S_{\{k+1:n\}}$
because sentences whose generation probabilities change with the removal of events in $S_k$ are limited to $S_{\{k+1:n\}}$ when using left-to-right LM, such as GPT-2.
In summary, we estimate the coherence score $c(S)$ as follows:
\begin{align}
    &c(S_{\{1:n\}}) := \frac{1}{\lvert S_{\{k+1:n\}} \rvert} \log P^{}_{}(S_{\{k+1:n\}} \mid S_{\{1:k-1\}}, S_{k})\text{,} \\
    &c(\widetilde{S}_{\{1:n\}}) := \frac{1}{\lvert S_{\{k+1:n\}} \rvert} \log P^{}_{}(S_{\{k+1:n\}} \mid S_{\{1:k-1\}}, r(S_k))
    \label{coherence-computation}
    \text{,}
\end{align}
where $\lvert S_{\{k+1:n\}}\rvert$ denotes the number of tokens in $S_{\{k+1:n\}}$.
In order to compute the salience score for each narrative text's last sentence, we add a special token that indicates the end of a text at the end of each narrative.
The proposed methods can compute the salience score for the last sentence using the generation probability of this special token.
See Appendix~\ref{app:approach} for further details.
\begin{wraptable}[8]{r}{0.5\textwidth}
    \centering
    \setlength{\tabcolsep}{5pt}  %
    \renewcommand{\arraystretch}{0.87} %
\input{tables/dataset_stats_modified.tex}
\caption{Statistics of the \propplearner{} corpus}
\label{table:dataset-stats}
\end{wraptable}

\section{Experiments}
\label{sec:experiments}
In this section, we provide empirical evidence that the proposed methods can evaluate event salience in narratives.
Concretely, we applied the proposed method with three event-removal methods on a manually annotated folktale dataset and confirmed their performance.
\subsection{Experimental setup}
\label{main-experiment}
\paragraph{Dataset}
\label{dataset:propplearner}
We used the \propplearner{} corpus~\cite{finlayson2015:ProppLearner}, which contains 15 Russian folktales. %
\begin{wraptable}[38]{r}{0.5\textwidth}
    \centering
    \small
    \setlength{\tabcolsep}{3pt}  %
    \renewcommand{\arraystretch}{1.4} %
    \input{tables/main_results_modified_vertically_long_v3.tex}
    \caption{MAP scores for the proposed methods and the baseline methods. We report the MAP score for random baseline method as the average over 10 seeds (standard deviation $=0.015$). Values with a dagger mark are statistically significant improvements over the random baseline method, which was tested using the Wilcoxon signed-rank test~\cite{wilcoxon1945} with $p<0.05$. The bold score is the best performance in our proposed methods alone. The bold italic score is the best performance in combination methods of our proposed methods and the TF-IDF baseline method.}
    \label{table:main_results}
\end{wraptable}
In the corpus, verbs corresponding to the Proppian function, i.e., salient event, which is the predecessor of CFs are annotated.
Following the task setup, our goal is to detect sentences that contain such verbs, i.e., salient events.
The \propplearner{} corpus includes POS and semantic role annotations, which are used by \verbanonym{} and \pasanonym{}.
Table~\ref{table:dataset-stats} shows the statistics of the \propplearner{} corpus.
\paragraph{Language model (fine-tuning)}
We used GPT-2 as a pre-trained language model for computing coherence scores%
\footnote{We used transformers~\cite{wolf2019arxiv:transformers} pre-trained model (12-layer, 768-hidden, 12-heads, 117M parameters).}.
Note that \newcite{see2019conll:massively} reported that GPT-2 outperforms state-of-the-art story generation models in coherence evaluation.
In Appendix~\ref{app:pre_exp}, we provide further evidence that GPT-2 can accurately evaluate the coherence of the narrative texts used in our experiments.
Moreover, we consider three fine-tuning settings.
\begin{enumerate}
    \item \textbf{No fine-tuning}
    \item Fine-tuning GPT-2 on \textbf{BookCorpus}~\cite{zhu2015:bookcorpus} as domain adaptation.
    \item Fine-tuning GPT-2 on \textbf{\propplearner{}} as transductive domain adaptation~\cite{vapnik1998:transductive,ouchi2019emnlp:transductive}.
\end{enumerate}
\paragraph{Baselines}
We compared the proposed methods with the following baseline methods:
\begin{itemize}
    \item Random baseline: This method assigns a random score in the range $[0, 1)$ to each sentence. 
    \item Sentence position baseline (ascending): This method assigns a score based on the position of each sentence. Here, we assumed that a sentence closer to the story's end has higher salience~\cite{friedland2008cikm:joke}.
    \item Sentence position baseline (descending): This method assigns a score in the opposite way to sentence position baseline (ascending).
    \item TF-IDF baseline: This method assigns the sum of the TF-IDF values\footnote{We used the score referred to as T3 in \newcite{nobatal2003:evaluation}} of the words in the sentence for each sentence.
\end{itemize}
\paragraph{Evaluation metric}
We cast salience estimation as a ranking problem following \newcite{liu2018emnlp:salience}, where each method ranks a sentence based on its salience score.
We used mean average precision (MAP) as an evaluation metric~\cite{manning2008:evaluationIR}.
We calculated the average precision for each story and reported their macro average score.
\subsection{Experimental results}
Table~\ref{table:main_results} shows the experimental results.
The results show all proposed methods consistently outperform the random baseline method, and the proposed method (\sentencedel{}, \propplearner{}) yields the best performance.
\paragraph{Event removal methods}
We found \sentencedel{} performed comparably to or relatively better than \verbanonym{} and \pasanonym{}~\footnote{We also tried other event removal methods, such as replacing verbs with common verbs and simultaneously replacing arguments with random input vectors; however, these modifications did not affect the results significantly.}.
We employed \verbanonym{} and \pasanonym{}, aiming to remove event information from the sentence more elaborately than \sentencedel{}.
However, experimental results show that these methods do not improve the proposed method.
We suspect unnatural sentences produced by the operations in \verbanonym{} and \pasanonym{} might negatively affect inference of the language model, indicating some room for improvement in how to remove events from a sentence.

\paragraph{Effect of fine-tuning GPT-2}
Fine-tuning GPT-2 on the BookCorpus slightly but consistently improved the proposed methods with \sentencedel{}, \verbanonym{} and \pasanonym{}.
We found that fine-tuning GPT-2 on the \propplearner{} corpus (transductive setting) also improved the proposed methods with \sentencedel{} and \pasanonym{}.
In addition, we found that our methods' MAP scores and LM's perplexity on the \propplearner{} corpus were strongly correlated. %
For each of \sentencedel{}, \verbanonym{}, and \pasanonym{}, the Spearman's rank correlation coefficient between the MAP score in three LM settings and the LM's perplexity were $-1.0$, $-0.5$, and $-1.0$.
This result shows that the better the LM fits the evaluation corpus, the better our methods perform.
\paragraph{Combining the proposed method and the baseline method}
We performed additional experiments with the same setting by combining each proposed method with the TF-IDF baseline method, which is the best baseline method.
We normalized salience scores of each proposed method and the TF-IDF baseline method to [0, 1] within each story~\footnote{We used \emph{scikit-learn}~\cite{sklearn_api} implementation of MinMaxScaler} and then added them to obtain the final salience score. 
Results are shown in Table~\ref{table:main_results} as \textbf{+TF-IDF}.
For all cases, combination methods consistently improved MAP scores more than our proposed methods alone or the TF-IDF baseline method alone.
The combination of the proposed method (\sentencedel{}, BookCorpus) and the TF-IDF baseline method and the combination of the proposed method (\pasanonym{}, \propplearner{}) and the TF-IDF baseline method achieved the best performance among all methods.
The Wilcoxon signed-rank test on the best combination method (i.e., combination of the proposed method (\sentencedel{}, BookCorpus) and the TF-IDF baseline method) and the TF-IDF Baseline method resulted in a p-value of $0.21$. 
This result suggests that TF-IDF-based salience cues are complementary to Barths' CFs-based cues, and they have been merged into a better measure of event salience.

Appendix~\ref{app:analysis} shows examples of salience evaluation results in toy \emph{Cinderella} story and qualitative analysis of the behavior of our proposed method.
\section{Discussion and future work}
\label{sec:discussion}
One promising direction for improving our proposed methods is to improve the narrative coherence evaluator.
For more accurate coherence evaluation, the coherence evaluator needs to have world knowledge and common sense reasoning skills.
Imagine the story of \emph{Cinderella}.
To be able to identify that the absence of event \textit{The prince falls in love with Cinderella} leads to coherence reduction, an ideal coherence evaluator needs to recognize that this event has a strong causal relation (in this case, precondition) with the next event \textit{Cinderella marries the prince}.
Recently, several techniques have been proposed to provide language models with more world knowledge~\cite{guan2020tacl:pretrained} and to enhance the common sense reasoning skills of language models~\cite{mao2019emnlp:improving}.
Evaluating the coherence of a narrative using these LMs can potentially improve our proposed methods.

\section{Conclusions}
Inspired by the Barthes' definition of cardinal functions in narratology, we have proposed methods to estimate event salience in a narrative in an unsupervised manner using an LM.
In our proposed methods, we have removed events from a narrative text and have estimated event salience by comparing the coherence score of the original narrative text with that of the event-removed narrative text.
Experiments on a folktales dataset have demonstrated that the proposed methods outperformed baseline methods and fine-tuning the LM on a narrative text is an effective way to improve the proposed methods.

\bibliographystyle{coling.bst}
\bibliography{references.bib}

\clearpage
\appendix
\section{Preliminary experiments}
\label{app:pre_exp}
In this section, we preliminary assess the ability of GPT-2 to evaluate the coherence of texts. 
Results support use of GPT-2 as a coherence evaluator in our methods.
\subsection{Preliminary experiment 1: Assessing GPT-2 as a coherence evaluator}
\label{pre-experiment1}
In this preliminary experiment, we evaluated GPT-2 in a sentence ordering task, which is a common task for evaluating discourse coherence models~\cite{barzilay2008cl:entitygrid,jurafsky2017emnlp:coherence}.
\newcite{see2019conll:massively} reported that GPT-2~\cite{radford2019:gpt2} better captures narrative text's coherence compared to the state-of-the-art model in story generation.
However, \newcite{see2019conll:massively} evaluated GPT-2 in the document ranking task, a slightly different task from sentence ordering task.
Thus, we examined GPT-2's ability as a coherence evaluator in the sentence ordering (i.e., the common task for evaluating discourse coherence models) and provided further evidence that GPT-2 could accurately evaluate the coherence of the narrative texts used in our experiments.

Given a pair from an original document and one of its permutations, the task is to assign a higher coherence score to original one.
In evaluation, GPT-2 predicted that the text with a higher likelihood was more coherent.
We report accuracy as the ratio of the model's correct predictions.

For each of 15 narratives in \propplearner{}, we generated 80 random permutations.
Then we obtained 1,200 pairs from original narrative texts and one of its permutations.
Results showed that GPT-2 achieved 100\% accuracy.
\newcite{jurafsky2017emnlp:coherence} reported 87.3\% accuracy on the same task~\footnote{They used randomly selected paragraphs from Wikipedia as an original document. Therefore, we cannot directly compare our results with theirs.}.
Our result supported the validity of using GPT-2's likelihood to compute narratives' coherence.
\subsection{Discussion: On evaluation of discourse coherence models}
In a common task to evaluate a discourse coherence model (e.g., sentence ordering), the model is given an original text and an artificially created incoherent text; it is then required to score the former with a higher coherence score.
In our preliminary experiments, we created incoherent texts by shuffling sentences as a common practice in sentence ordering task, and \newcite{see2019conll:massively} created an  incoherent text by swapping adjacent sentences. 
These experiments demonstrated that GPT-2 can accurately perform these tasks.

However, as mentioned in \newcite{lai2018sigdial:discourse}, identifying a document’s original sentence order is not the same as distinguishing low and high coherence.
Just identifying sentences' correct order is not sufficient for evaluating coherence models, and we believe that more elaborate evaluation methods are needed.
\newcite{lai2018sigdial:discourse} provides a dataset that addresses this issue, but does not include the texts in the narrative domain.
\subsection{Preliminary experiment 2: Sanity check via sentence deletion detection}
\label{pre-experiment2}
In this preliminary experiment, we validated whether our method could detect event elimination as a step prior to identifying event salience.
Given a narrative comprising $n$ sentences $S_{\{1:n\}} := \{S_1,\dots,S_n\}$, in which every sentence can be regarded as highly salient, and the target sentence $S_k \in S_{\{1:n\}}$, we evaluated whether GPT-2 can detect the sentence's deletion as a reduction in the subsequent story's likelihood: $\sigma(S_k,S_{\{1:n\}}) > 0$.
If our method could not do so, our methods would be unlikely to work because they are required to reduce the subsequent story's likelihood when the target sentence (to be removed) contains salient event.
\paragraph{Dataset}
For this experiment, we need a dataset that allows us to assume that every sentence in a story is highly salient (i.e., removing any sentence would result in a significantly incoherent narrative).
We used ROCStories~\cite{mostafazadeh2016naacl:roc} because it is designed to meet the requirement that each story captures a rich set of causal and temporal common sense relations among daily events.
Each story contains five sentences. %
As the event removing method, we examine \sentencedel{} in this preliminary experiment.
We used the 2016 Spring Set (45,495 stories) and the 2017 Winter Set (52,664 stories). 
\paragraph{Task Setting}
We calculated accuracy as the percentage of cases in which sentence deletion was correctly detected as $\sigma(S_k,S_{\{1:n\}}) > 0$.
Random prediction would result in 50\% accuracy.
\paragraph{Result}
\sentencedel{} with GPT-2 (No-fine-tuning) achieved 94\% accuracy in both the 2016 Spring Set and the 2017 Winter Set.
This result shows that \sentencedel{} can detect event deletion with \sentencedel{} as a reduction in the subsequent story's likelihood . %

\section{Details of proposed approach}
\label{app:approach}
GPT-2, which is an LM we used for computing coherence score, has a limitation of input length and we can't always input the entire narrative text.
Thus we practically compute coherence score $c(S_{\{1:n\}})$ as follows:
\begin{align}
    c(S) = \frac{1}{\lvert S_{\{k+1:n-\ell'\}} \rvert} \log P^{}_{}(S_{\{k+1:n-\ell'\}} \mid S_{\{1+\ell:k-1\}}, S_{k}) \text{,}
\end{align}
where $\lvert S_{\{k+1:n-\ell'\}} \rvert$ denotes the number of tokens in $S_{\{k+1:n-\ell'\}}$, so as $c(\widetilde{S})$.
$P(S_i)$ is computed as the product of the probability of words: 
\begin{align}
    & \log P(S_i | \text{context})
    = \log P(w^{(i)}_1 | \text{context}) \nonumber\\
    & + \sum_{j=2}^{|S_i|} \log P(w^{(i)}_{j}|\text{context}, w^{(i)}_1, \dots, w^{(i)}_{j-1})
\text{.}
\end{align}
$\lvert S_{\{i:j\}}\rvert$ denotes the sum of the number of words in $(S_i,\dots,S_j)$.
$\ell'$ and $\ell$ are thresholds determined by input length limitation of the language model, $L$.
We determine $\ell'$ and $\ell$ so that $\lvert S_{\{k+1:n-\ell'\}} \rvert + \lvert S_{\{1+\ell:k\}} \rvert$ are less than or equal to $L$ and have a maximum value, respectively.

As mentioned at the end of Section\ref{modeling-event-salience}, we add a special token that indicates the end of a text at the end of each narrative text for computing the salience score for the last sentence in each narrative text~\footnote{We used \textless{}\textbar{}endoftext\textbar{}\textgreater{} special token in transformers~\cite{wolf2019arxiv:transformers} implementation.}.
The generation probability of this special token is used only when computing the salience score for the last sentence, otherwise it is ignored.
\section{Estimating event salience for toy example}
\label{app:analysis}
\begin{table*}[h]
    \centering
    \input{tables/toy_example_modified.tex}
    \caption{The behavior of the proposed method (\sentencedel{}, No fine-tuning) in toy \textit{Cinderella} story. Our method gives sentences a high salience score if the target sentence contains a salient event.}
    \label{table:toy-example}
\end{table*}
Table~\ref{table:toy-example} shows the behavior of the proposed method (\sentencedel{}, No fine-tuning) in toy example, \textit{Cinderella}.
We found the last sentence tended to have a large variance in its salience score because only one special token in the succeeding story is used for estimating saliency.
\newpage
\begin{table*}[h]
    \centering
    \small
    \setlength{\tabcolsep}{4pt}  %
    \renewcommand{\arraystretch}{1.7} %
    \input{tables/model_behavior_modified.tex}
    \caption{The more detailed behavior of our proposed method (\sentencedel{}, No fine-tuning) in toy \textit{Cinderella} story. The value in row $i$, column $j$, represents the difference in $S_i$'s generation probability (token-wise likelihoods are averaged within a sentence) before and after $S_j$ is removed from the story. A Large value indicates that the removal of $S_j$ greatly reduces the generation probability of $S_i$.}
    \label{table:model_behavior}
\end{table*}
Table\ref{table:model_behavior} shows the more detailed behavior of the proposed method (\sentencedel{}, No fine-tuning) in \textit{Cinderella}.
For example, the last row shows that deleting salient sentences (e.g., $S_3, S_5$) resulted in a larger decrease in the likelihood of the ending sentences ($S_6$) than deleting less salient sentences (e.g., $S_1, S_4$).
In addition, if we look at the likelihood difference of $S_3$, \textit{Cinderella goes to the ball}, the likelihood dropped more when the $S_2$, related sentence to $S_3$, is removed than when $S_1$, which is unrelated to $S_3$ is removed.
\end{document}

%% file: tables/dataset_stats_modified.tex
\begin{tabular}{lr}
\toprule
\# of stories                    & 15     \\
\# of sentences                  & 1302   \\
\# of words                      & 18862  \\
\# of functions (salient events) & 170    \\
average \# of sentences / story  & 86.8   \\
average \# of words / story      & 1257.5 \\
average \# of functions / story  & 11.3   \\ \bottomrule
\end{tabular}

%% file: tables/main_results_modified_vertically_long_v3.tex
\begin{tabular}{llll}
\toprule
Method                         & + TF-IDF           & Fine-tuning  & MAP                     \\ \midrule
Random                         & -                  & -            & 0.213                   \\
Sentence position (asc)  & -                  & -            & 0.277$^\dagger$                   \\
Sentence position (desc) & -                  & -            & 0.185                   \\
TF-IDF                         & -                  & -            & 0.279$^\dagger$                   \\ \midrule
\multirow{6}{*}{Proposed method w/ \textbf{SD}}            & \multirow{3}{*}{-} & -            & 0.261$^\dagger$                   \\
                               &                    & BookCorpus   & 0.265$^\dagger$                   \\
                               &                    & ProppLearner & \textbf{0.280}$^\dagger$          \\ \cmidrule{2-4} 
                               & \multirow{3}{*}{\ding{51}} & -            & 0.294$^\dagger$                   \\
                               &                    & BookCorpus   & \textit{\textbf{0.301}}$^\dagger$ \\
                               &                    & ProppLearner & 0.295$^\dagger$                   \\ \midrule
\multirow{6}{*}{Proposed method w/ \textbf{VA}}            & \multirow{3}{*}{-} & -            & 0.245                   \\
                               &                    & BookCorpus   & 0.258$^\dagger$                   \\
                               &                    & ProppLearner & 0.219                   \\ \cmidrule{2-4} 
                               & \multirow{3}{*}{\ding{51}} & -            & 0.286$^\dagger$                   \\
                               &                    & BookCorpus   & 0.287$^\dagger$                   \\
                               &                    & ProppLearner & 0.266$^\dagger$                   \\ \midrule
\multirow{6}{*}{Proposed method w/ \textbf{PAA}}           & \multirow{3}{*}{-} & -            & 0.254$^\dagger$                   \\
                               &                    & BookCorpus   & 0.258$^\dagger$                   \\
                               &                    & ProppLearner & 0.266                   \\ \cmidrule{2-4} 
                               & \multirow{3}{*}{\ding{51}} & -            & 0.285$^\dagger$                   \\
                               &                    & BookCorpus   & 0.295$^\dagger$                   \\
                               &                    & ProppLearner & \textit{\textbf{0.301}}$^\dagger$ \\ \bottomrule
\end{tabular}

%% file: tables/toy_example_modified.tex
\begin{tabular}{cllr}
\toprule
Including salient event & \multicolumn{2}{c}{Sentence}                                                                                                               & Salience score\\ \midrule
-                       & $S_1$ & Cinderella draws water from a well.                                                                                          & $0.193$    \\
 \ding{51}                      & $S_2$ & \begin{tabular}[c]{@{}l@{}}A fairy godmother appears and provides Cinderella\\ with clothes, a carriage, and a coachman.\end{tabular} & $0.309$    \\
\ding{51}                       & $S_3$ & Cinderella goes to the ball.                                                                                                           & $0.214$    \\
-                       & $S_4$ & \begin{tabular}[c]{@{}l@{}}Cinderella greets her stepsisters at the venue\\ , but they do not notice.\end{tabular}                     & $-0.014$    \\
\ding{51}                       & $S_5$ & The prince falls in love with Cinderella.                   & $0.394$    \\
\ding{51}                       & $S_6$ & Cinderella marries the prince.                                                                                                         & $-0.112$   \\ \bottomrule
\end{tabular}

%% file: tables/model_behavior_modified.tex
\begin{tabular}{cllcccccc}
\toprule
              &             &                                                                                                                                        & \multicolumn{6}{c}{likelihood diff when deleting}                                                                                                                      \\ \cmidrule{4-9} 
salient event &             & sentence                                                                                                                               & $S_1$    & \cellcolor[HTML]{EFEFEF}\textbf{$S_2$} & \cellcolor[HTML]{EFEFEF}\textbf{$S_3$} & $S_4$     & \cellcolor[HTML]{EFEFEF}\textbf{$S_5$} & \cellcolor[HTML]{EFEFEF}\textbf{$S_6$} \\ \midrule
              & $S_1$          & Cinderella draws water from a well.                                                                                                    & -     & \cellcolor[HTML]{EFEFEF}-           & \cellcolor[HTML]{EFEFEF}-           & -      & \cellcolor[HTML]{EFEFEF}-           & \cellcolor[HTML]{EFEFEF}-           \\ \midrule
\ding{51}             & \textbf{$S_2$} & \begin{tabular}[c]{@{}l@{}}A fairy godmother appears and provides Cinderella\\  with clothes, a carriage, and a coachman.\end{tabular} & $0.562$ & \cellcolor[HTML]{EFEFEF}-           & \cellcolor[HTML]{EFEFEF}-           & -      & \cellcolor[HTML]{EFEFEF}-           & \cellcolor[HTML]{EFEFEF}-           \\ \midrule
\ding{51}             & \textbf{$S_3$} & Cinderella goes to the ball.                                                                                                           & $0.807$ & \cellcolor[HTML]{EFEFEF}$0.876$       & \cellcolor[HTML]{EFEFEF}-           & -      & \cellcolor[HTML]{EFEFEF}-           & \cellcolor[HTML]{EFEFEF}-           \\ \midrule
              & $S_4$          & \begin{tabular}[c]{@{}l@{}}Cinderella greets her stepsisters at the venue\\ , but they do not notice.\end{tabular}                     & $0.087$ & \cellcolor[HTML]{EFEFEF}$0.257$       & \cellcolor[HTML]{EFEFEF}$0.296$       & -      & \cellcolor[HTML]{EFEFEF}-           & \cellcolor[HTML]{EFEFEF}-           \\ \midrule
\ding{51}             & \textbf{$S_5$} & The prince falls in love with Cinderella.                                                                                              & $0.175$ & \cellcolor[HTML]{EFEFEF}$0.274$       & \cellcolor[HTML]{EFEFEF}$0.034$       & $-0.111$ & \cellcolor[HTML]{EFEFEF}-           & \cellcolor[HTML]{EFEFEF}-           \\ \midrule
\ding{51}             & \textbf{$S_6$} & Cinderella marries the prince.                                                                                                         & $0.139$ & \cellcolor[HTML]{EFEFEF}$-0.113$      & \cellcolor[HTML]{EFEFEF}$0.220$       & $0.082$  & \cellcolor[HTML]{EFEFEF}$0.394$       & \cellcolor[HTML]{EFEFEF}-           \\ \bottomrule
\end{tabular}